\documentclass[letterpaper, 10 pt, conference]{ieeeconf}  

\IEEEoverridecommandlockouts                              
\usepackage{graphicx}
\usepackage[]{algorithm2e}
\graphicspath{ {./} }

\title{\LARGE \bf
A Computational Framework for Motor Skill Learning
}

\author{Krishn Bera$^{1}$ Tejas Savalia$^{2}$ Bapi Raju$^{3}$
\thanks{}
\thanks{$^{1}$ International Institute of Information Technology, Hyderabad
        {\tt\small krishn.bera@research.iiit.ac.in}}%
\thanks{$^{2}$ International Institute of Information Technology, Hyderabad
        {\tt\small 	savalia.tejas@research.iiit.ac.in}}%
\thanks{$^{3}$ International Institute of Information Technology, Hyderabad and University of Hyderabad
        {\tt\small raju.bapi@iiit.ac.in}}%
}

\begin{document}

\maketitle
\thispagestyle{empty}
\pagestyle{empty}

\begin{abstract}

There have been numerous attempts in explaining the general learning behaviours by various cognitive models. Multiple hypotheses have been put further to qualitatively argue the best-fit model for motor skill acquisition task and its variations. In this context, for a discrete sequence production (DSP) task, one of the most insightful models is Verwey’s Dual Processor Model (DPM). It largely explains the learning and behavioural phenomenon of skilled discrete key-press sequences without providing any concrete computational basis of reinforcement. Therefore, we propose a quantitative explanation for Verwey’s DPM hypothesis by experimentally establishing a general computational framework for motor skill learning. We attempt combining the qualitative and quantitative theories based on a best-fit model of the experimental simulations of variations of dual processor models. The fundamental premise of sequential decision making for skill learning is based on interacting model-based (MB) and model-free (MF) reinforcement learning (RL) processes. Our unifying framework shows the proposed idea agrees well to Verwey’s DPM and Fitts’ three phases of skill learning. The accuracy of our model can further be validated by its statistical fit with the human-generated data on simple environment tasks like the grid-world.

\end{abstract}

\section{INTRODUCTION}

A wide range of our day to day activities like walking, riding a bicycle or writing involve our seemingly ‘innate’ ability to acquire and execute motor skills. It is for this reason that various domains in cognitive science have seen an ample amount of research in motor skill acquisition. The experiments have tried coming up with various cognitive models to best explain the motor skills phenomenon by hypothesizing various theories of representation, learning and execution. The procedural skill and habitual learning are a kind of non-declarative or implicit memory in our brain.
\\
\\
Various tasks like the pursuit rotor task \cite{c1}, the m × n task \cite{c2}, serial reaction time (SRT) task \cite{c3} and discrete sequence production (DSP) task \cite{c4} among others have been used to study motor skill learning. However, we are specifically looking at the DSP task because it builds upon the notion that sequential control occurs at the cognitive level as well as autonomous motor level reflecting an aspiration for a much broader application. Verwey also proposes a Dual Processor Model (DPM) based on two parallel processor architecture to explain the DSP task learning. We will see that the DPM not only explains the DSP phenomenon quite well but also relates to our interacting model-based (MB) and model-free (MF) reinforcement learning (RL) framework for skill learning. 
\\
\\
Verwey also proposes a three-phase execution model on the dual processor architecture. He classifies these phases similar to Fitts’ and Wulf’s categories \cite{c5} \cite{c6} - namely into cognitive, associative and motor phase. The cognitive phase is a verbal stage where the movements are slow, inconsistent and inefficient requiring a significant cognitive ability. A substantial part of the movement is controlled consciously and deliberately. The second phase represents the associative phase where movements become more fluid, reliable and efficient while requiring less cognitive activity. However, some parts of the movement are controlled consciously while others seem to be automatically controlled. The latest phase represents the autonomous motor phase where the movements are accurate, consistent and efficient. Since the movement is largely controlled automatically, little or no cognitive activity is required.

\section{METHOD}

Our method represents a computational approach to the skill learning task by proposing a reinforcement learning model for sequential decision making. A reinforcement learning model learns to reach a goal by maximizing its rewards or utilities by interacting with an environment rather than being explicitly taught. A reinforcement learning agent that learns from the consequences of its actions tries to optimize between its exploitation-exploration policies. Our proposed RL paradigm for sequential decision making is reducible to sequential skill learning because the former involves selecting a sequence of actions to accomplish a goal, or to maximize an RL cost function. Our model represents a reinforcement-maximizing variation of sequential decision making, that is, given $ s_i, s_{i+1}, ... $, we want to choose an action $a_j$ at time step $j$ that will lead to receiving maximum total reinforcement in the future. \cite{c7}
\\
\\
Verwey’s DPM for DSP is a qualitative theory of sequential skill learning without any quantitative treatment. The DPM suggests that the initial phase of training is dominated by the cognitive processor whereas the later phase of the training is dominated by the motor processor. Verwey argues that ‘reaction’ mode of execution is associated with the slow learning phase. It is dominated by the closed-loop control of the cognitive processor. After substantial practice, the execution switches to the ‘chunking’ mode where sequence learning happens without stimuli in an open-looped control of the motor processor. He also proposes ‘associative’ mode of execution with intermediate execution speed due to successive priming of sequences.
\\
\\
Therefore, we try to build up a quantitative computational framework based on interactive model-free and model-based reinforcement learning paradigm that supports the observations and processes of Verwey’s DPM for DSP. Our proposed and generic model can be extended to more general motor skill learning tasks. Our effort is to combine the qualitative and quantitative theories based on a best-fit model of the experimental simulations of various dual processor models. As a high-level abstract, our proposed idea is that the early phase of training is dominated by a cognitive model-based reinforcement learning process whereas, after substantial practice in the later stage, the motor processor facilitated model-free reinforcement learning is dominant. Some of the approaches that can be potential candidates for such a computational paradigm are the weighted stream model and parallel stream model of reinforcement learning. In case of the weighted stream model of model-based and model-free learning, the output depends upon the weighted summation of both streams and the dominant stream would contribute to the decision. In the parallel stream model, there are competing and parallel streams of MB and MF learning.
\\
\\
We tried testing the framework proposed in \cite{c8}. In this computational work, they present a combination of reinforcement learning mechanisms applied to a toy grid-world environment and further argue that this representation of the combination is better than individual entities of the dichotomy. It is further hypothesized that such a framework can then be generally extended to explain skill acquisition, investigation of which remains. For modelling skill learning, they combine them with initial trials using model-based and later trials using model-free reinforcement learning. The model-based RL is implemented by a simple Depth-Limited Search (DLS) and model-free RL by Temporal Difference (TD) learning. Both the processes share a common look-up table while searching for action with best expected reward. The two processes interact hierarchically across trials -- the first trial being completely model-based. The next trial consists of a model-based DLS at the first position in the grid world and model-free TD at the next; alternating so on. 

\begin{algorithm}
Initialize Internal Model with equal transition probabilities and zero value for each state\;
\For{Iteration i}
{
  Agent at the start state\;
  j = 0\;
  \While{Till reward reached}
  {
    \eIf{j mod (i/factor) == 0 or j mod chunk size == 0}
    {
    	Take Model-Based step:\;
        Build a search tree from the current state based on the internal model of the environment.\;
        Take an action that leads to the state maximizes the expected reward at the end of the search tree.\;
        Update Transition Probabilities\;
        Update Values of the state and Q-values of the actions from that state\;
    }
    {
    	Take Model-Free Step:\;
        Take the action that maximizes the corresponding Q-value.\;
        Update Values of the State and Q values of the corresponding action\;
    }
  }
}
\caption{The Dual Process algorithm in \cite{c8}}
\end{algorithm}
.
\\

Many other related works have tried bringing into light the working of dual processors for behavioural choices. These models try proposing different mechanisms through which the model-based and model-free processes interact. In spite of all the working differences, all of them confirm the dual-processor model of model-based and model-free reinforcement learning for sequential decision making. This large corpus of work will form the fundamental inspiration on which we build up our computational framework. Discussed here are a few of the approaches towards framing a quantitative reinforcement learning framework of the dual processor model.
\\
\\
The fMRI results from the experiment in \cite{c10} confirms the presence of dual learning processes involved in guiding the sequential decision making. The neural signatures of a model-free process are found in dorsolateral striatum (DLS) whereas that of a model-based process is found in pre-frontal cortex (PFC) areas of the brain. The experiment proposed in \cite{c9} explores how the PFC and DLS dual systems arbitrate control for behavioural choices. Using computational theory of reinforcement learning, they suggest an uncertainty based Bayesian principle of arbitration for the competing parallel processes. The more dominant system is the one which predicts actions with least uncertainty and most accuracy. Temporal Difference Model (TDM) is yet another framework that is recently proposed in \cite{c11}. It proposes a hybrid model-based and model-free reinforcement learning algorithm. In this framework, initially, the planning of action sequences is executed by a model-based process using abstract sub-goals to the target state. Overtime, each sub-goal is learned by a trial-and-error approach much along the lines of a model-free process. The proposed Temporal Difference Model (TDM) is where there is a goal-conditioned subdivision of the action sequences. For an optimal state-action-reward policy, model-based learning guides the model-free learning at periodic intervals that ultimately leads to the goal state.
\\
\\
We propose to test the above-mentioned models by simulating simple environments like the grid-world. We can compare our findings with the general DPM observations from the DSP task. The accuracy of our model can further be validated with the actual human subjects data on simple environment tasks like the grid-world.

\section{RESULTS}

\begin{figure}[thpb]
\centering
\includegraphics[width=\linewidth]{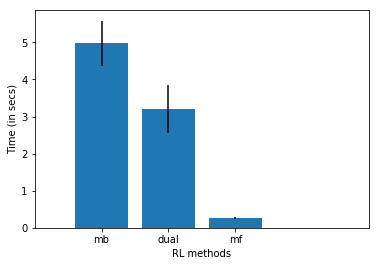}
\caption{MB, MF and Dual comparison: Time taken to reach the goal state }
\label{fig:time}
\end{figure}

\begin{figure}[thpb]
\centering
\includegraphics[width=\linewidth]{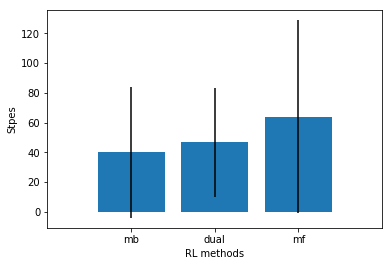}
\caption{MB, MF and Dual comparison: Steps taken to reach the goal state }
\label{fig:steps}
\end{figure}

\begin{figure}[thpb]
\centering
\includegraphics[width=\linewidth]{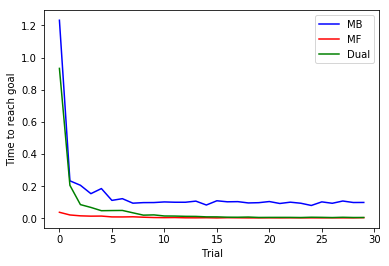}
\caption{MB, MF and Dual comparison: Transveral time with practice }
\label{fig:graph}
\end{figure}

\begin{figure}[thpb]
\centering
\includegraphics[width=\linewidth]{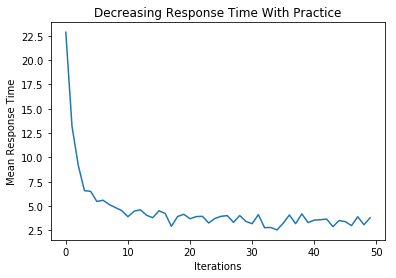}
\caption{ Response time decreases with practice }
\label{fig:response_time}
\end{figure}

In the typical DSP task, the S-R outputs are slow at first due to the associative phase of execution where the cognitive processor is dominant. As evident from the Fig. \ref{fig:time} and \ref{fig:response_time}, the average response time in the initial training phase is reasonably high. This is mostly attributed to the model-based depth-limited search. The model-based RL steps routinely take more time because they are data-efficient especially when there is not much information available about the environment at the start of the experiment. However, a purely model-based learning will follow the minimal number of steps to reach the goal state provided the optimal actions it takes by calculating the state prediction error. See Fig. \ref{fig:steps}. Therefore, the model-based reinforcement learning can be taken to obtain a typical representation of the early phase practice in sequential motor skill learning task like DSP.
\\
\\
However, in the later phase of the DSP task, the execution speed gradually increases to attain a maximum after substantial practice. This is because the actions are now executing in chunking mode which is autonomously controlled by the motor processor. We see a minimal cognitive processor activity in this phase as the motor processor executes a sequential learning task in an open-looped control. The motor processing takes a significantly lesser execution time at each step because it uses experience directly in form of reward prediction error. See Fig. \ref{fig:time}, \ref{fig:response_time}. During the later phase of the training, we see in Fig. \ref{fig:graph} that the mean response time decreases to strike a constant time with little or no fluctuations. This is indicative of the fact that the model-free process has taken over from the earlier model-based reinforcement learning process. However, it is critical to note that a purely model-free process will demand a significantly more extensive time to converge at a goal state as evident from the number of steps taken by it in Fig \ref{fig:steps}.

\section{DISCUSSIONS}

\begin{figure}
  \includegraphics[width=\linewidth]{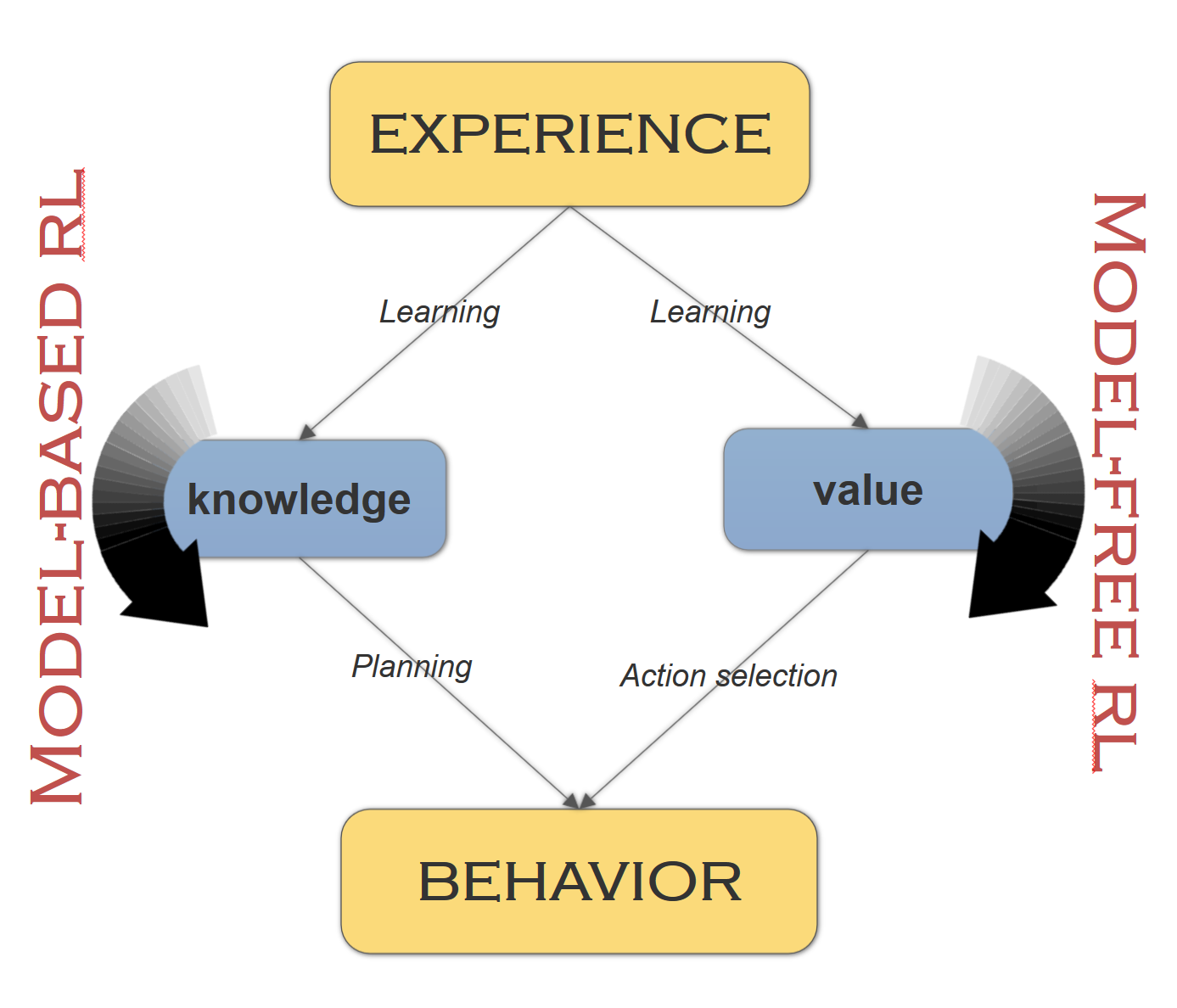}
  \caption{Model-free and model-based Reinforcement Learning. Ref. Toussiant (2010) }
  \label{fig:overview}
\end{figure}

From the foregoing analysis of the results, we see that the procedural and habitual memory for skill learning can be well modelled by interacting model-based and model-free reinforcement learning paradigm. In a significant measure of performance, the model proposed in \cite{c8} can be understood as a quantitative interpretation of the Verwey’s qualitative theory of sequential motor skill acquisition.
\\
\\
The framework also correlates surprisingly well with Fitts’ three phases of execution. The initial cognitive phase is dominated by model-based processes whereas the late motor phase is controlled by model-free processes. The intermediate associative phase seems to represent an outcome of the interacting model-free and model-based processes. Therefore, such a reinforcement schema provides us with a computational outline of the sequential motor skill learning task. Figure \ref{fig:overview} shows an overview of interacting MB-MF reinforcement learning paradigm.
\\
\\
A scope of future work lies in the determination of intricacies of the unanswered questions like if the model-based and model-free reinforcement is competitively racing against each other or if it is the weighted sum of their outputs that need to be calculated for action reward prediction. Moreover, other algorithmic details need to be figured out like - if model-based and model-free processes have a common or individual reward table, or say where is the exact trade-off between model-free steps and model-based steps in case of temporal difference models (TDM). From our initial results, it is evident that a combination of model-based and model-free reinforcement learning presents a functional account of the phases of skill learning. Such a framework would also enable us to obtain better insights into the other dichotomies of skill learning - habitual vs. goal-directed learning or implicit or explicit learning.

\section{CONCLUSIONS}

To conclude, we propose a unifying framework for sequential motor skill acquisition by providing a concrete quantitative underpinning to Verwey’s DPM. The major insight for such problems is that the interacting model-based and model-free reinforcement learning paradigms can be used to explain the typical behaviour and observations in sequential motor skill tasks. Our effort in bringing the qualitative and quantitative theory accomplishes the same. Moreover, the proposed computational framework is quite generic and can be potentially extended to general motor skill learning tasks in the future. 

\addtolength{\textheight}{-12cm}   



\end{document}